\pdfoutput=1

\documentclass[11pt]{article}

\usepackage{acl}

\usepackage{times}
\usepackage{latexsym}

\usepackage[T1]{fontenc}

\usepackage[utf8]{inputenc}

\usepackage{microtype}

\usepackage{amsmath} 
\usepackage{CJKutf8}
\usepackage{booktabs}
\usepackage{makecell}
\usepackage{graphicx}
\usepackage{multirow}
\usepackage{amssymb}

\title{Zero-Shot Cross-lingual Semantic Parsing}
\author{Tom Sherborne \and Mirella Lapata\\
Institute for Language, Cognition and Computation\\
  School of Informatics, University of Edinburgh \\
10 Crichton  Street, Edinburgh EH8 9AB\\
  \texttt{tom.sherborne@ed.ac.uk,~mlap@inf.ed.ac.uk} \\
}

\date{}

\begin{document}
\setlength{\abovedisplayskip}{2pt}
\setlength{\belowdisplayskip}{2pt}

\maketitle
\begin{abstract}
  Recent work in cross-lingual semantic parsing has successfully
  applied machine translation to localize parsers to new languages.
  However, these advances assume access to high-quality machine
  translation systems and word alignment tools.  We remove these
  assumptions and study cross-lingual semantic parsing as a zero-shot
  problem, without parallel data (i.e.,~utterance-logical form pairs)
  for new languages. 
  We propose a multi-task encoder-decoder model to transfer parsing
  knowledge to additional languages using only English-\mbox{logical form}
  paired data and in-domain natural language corpora in each new
  language.  Our model encourages language-agnostic encodings by
  jointly optimizing for logical-form generation with auxiliary
  objectives designed for cross-lingual latent representation
  alignment.
  Our parser performs significantly above translation-based baselines
  and, in some cases, competes with the supervised upper-bound.\footnote{Our code and data are available at {\tt
      github.com/tomsherborne/zx-parse}.} 
\end{abstract}



\section{Introduction}

Executable semantic parsing maps a natural language \emph{utterance}
to a \emph{logical form} (LF) for execution in some \emph{knowledge base}
to return a \emph{denotation}. The parsing task renders an utterance
as a semantically identical, but machine-interpretable, expression
\emph{grounded} in a denotation. The transduction between natural and
formal languages has allowed semantic parsers to become critical
infrastructure in building human-computer interfaces for question
answering,
\citep{SEMPRE-berant2013freebase,Liang:2016,kollar-etal-2018-alexa},
dialog systems \cite{artzi-zettlemoyer-2011-bootstrapping}, and
robotics \cite{dukes-2014-semeval}.

Recent advances in semantic parsing have improved accuracy for
neural parsers \citep{jia-liang-2016-data,dong-lapata-2016-language,wang-etal-2020-rat}
and examined their generalization capabilities with new dataset
challenges \citep{zhongSeq2SQL2017-wikisql,yu-etal-2018-spider}, in addition
to considering languages other than English
(\citealp{multilingsp-and-code-switching-Duong2017}; {\it inter
  alia.}). Prior work largely assumes that
utterance-logical form training data is parallel in all languages
\citep{multilingual-sp-hierch-tree-Jie2014}, or must be created with
human translation \citep{arch-for-neural-multisp-Susanto2017}. This
entry barrier to localization for new languages has motivated
the exploration of machine translation (MT) as an economical
alternative \citep{sherborne-etal-2020-bootstrapping,
  moradshahi-etal-2020-localizing}. However, MT can
introduce performance-limiting artifacts and struggle to accurately
model native speakers
\citep{riley-etal-2020-translationese}. Additionally, high-quality
machine translation is less viable for lower resource languages,
further limiting the appeal of MT-based approaches.

\begin{figure}[t]
    \centering
    \includegraphics[width=\columnwidth]{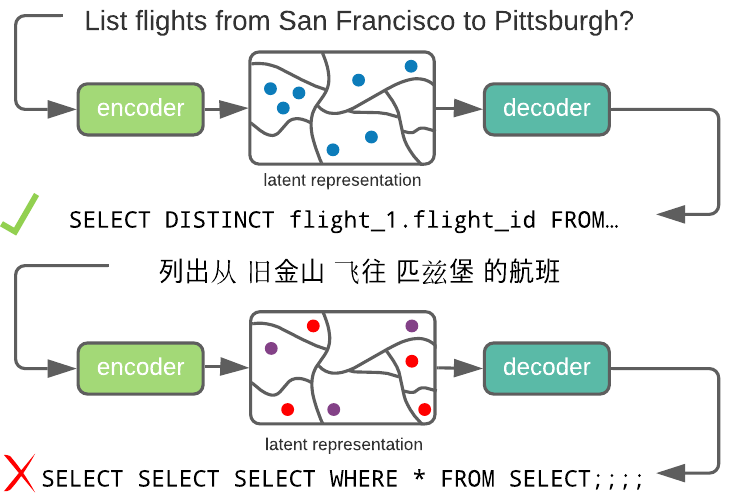}
    \caption{Accurate cross-lingual semantic parsing requires
      alignment of latent semantic representations across
      languages. The encoder generates a representation of the English
      utterance ({\color{blue} blue} points) to condition upon during
      decoding. Producing the same logical form from the equivalent
      Chinese utterance requires a similar encoding.  However, without
      alignment, the representation may partially match
      ({\color{violet} purple} points) or not at all ({\color{red}
        red} points), leading the decoder to generate an inaccurate,
      ill-formed query.  }
    \label{fig:problem}
    \vspace{-1.5em} 
\end{figure}

In this work, we propose a new approach for \textbf{zero-shot
  executable semantic parsing}.  Our method maximizes the success of
cross-lingual transfer for a parser, trained on English paired data
(\mbox{${\rm EN}\rightarrow{\rm LF}$}), to accurately generate logical
forms from new languages (${\rm X}\rightarrow{\rm LF}$).
Our goal is to parse utterances in a new language, $l$, without observing paired
training data for this language, suitable machine translation, or
bilingual dictionaries between~$l$ and English. Our critical
dependencies are a pre-trained language model and utterance-logical
form paired data for a source language (i.e.,~English). Aside from the
\emph{zero-shot} problem which is hard on its own (since paired data
is not available for new languages), our semantic parsing
challenge is further compounded with the difficulties inherent to
\emph{structured prediction} and the deficiency of copying strategies
 without gold token-level alignment \cite{zhu-etal-2020-dont-parse-insert}.

We conceptualize cross-lingual semantic parsing as a \emph{latent
  representation alignment} problem. As illustrated in
Figure~\ref{fig:problem}, we wish to encode different languages to an
overlapping latent space for the decoder to have any chance at
generating accurate logical forms.  To achieve this, we train a
decoder, conditioned upon encodings from a source language
(e.g.,~English), to generate logical forms \emph{and simultaneously}
train encodings of a new language (e.g., Chinese) to be maximally
similar to English.  We hypothesize that if latent representations are
aligned from a language-agnostic encoder, one can generate accurate 
logical forms from a new language without semantic parsing
training data and thus eliminate the errors outlined in 
Figure~\ref{fig:problem}.

Our approach adopts a multi-task learning paradigm and trains a
parser with auxiliary objectives, optimized to converge
representations of additional new languages. We encourage
language-agnostic representations by jointly optimizing for generating
logical forms, reconstructing natural language, and promoting language
invariance. Our intuition is that auxiliary losses can be
exploited to induce similarity in a multi-lingual latent space. The
effect of such alignment is that a decoder, trained only on
English, can recognize an encoding from another language and generate
the relevant logical form.  Similar multi-task approaches have been
successful in spoken-language understanding
\citep{van-der-goot-etal-2021-masked}, text simplification
\citep{mallinson-etal-2020-zero-shot-sent-simp,  Zhao_Chen_Chen_Yu_2020}, dependency parsing \citep{ahmad-etal-2019-cross}, and machine translation
\citep{DBLP:journals/corr/abs-1903-07091-Arivazhagan}. This work, to
our knowledge is the first attempt to devise auxiliary objectives for
executable semantic parsing as a zero-shot task. Our framework and hypothesis
are also sufficiently flexible for application in additional zero-shot
sequence transduction tasks.

Our motivation is to improve parsing for non-English languages with
maximal resource efficiency and minimal external dependencies beyond
native-speaker utterances. We, therefore, induce a shared multilingual
space without resorting to machine translation
\citep{sherborne-etal-2020-bootstrapping,
  moradshahi-etal-2020-localizing} and argue that our approach is
superior because it (a)~nullifies the introduction of translation or
word alignment errors and (b)~scales to low-resource languages without
reliable MT. Experimental results on Overnight
\cite{Overnight-Wang15,sherborne-etal-2020-bootstrapping} and a new
executable version of MultiATIS++ show that our parser generates
more accurate logical forms with a minimized cross-lingual transfer penalty from
English to French (FR), Portuguese (PT), Spanish (ES), German (DE),
Chinese (ZH), Hindi (HI), and Turkish (TR).  

\section{Related Work}
\label{sec:rw}
\paragraph{Cross-lingual Modeling} This area has recently gained
increased interest across several natural language understanding
settings \citep{zhao2020closer,nooralahzadeh-etal-2020-zero} with
benchmarks such as XGLUE \citep{liang-etal-2020-xglue} and XTREME
\citep{hu2020xtreme} allowing to study classification and generation
tasks for multiple languages.  Cross-lingual approaches have also been
developed for dependency parsing
\citep{tiedemann-etal-2014-treebank-xl-depparse,schuster-etal-2019-cross},
sentence simplification
\citep{mallinson-etal-2020-zero-shot-sent-simp}, and spoken-language
understanding (SLU;~\citealp{he2013multi-style};
\citealp{Upadhyay2018-multiatis}).

Pre-training has shown to be widely beneficial for a wide range of
cross-lingual models \citep{devlin-etal-2019-BERT,
  conneau-etal-2020-unsupervised-xlmr}.  By virtue of being trained on
massive corpora, these models purportedly learn an overlapping
cross-lingual latent space \citep{conneau-etal-2020-emerging} but have
also been identified as under-trained for some tasks
\citep{li-etal-2021-mtop}, shown poor zero-shot performance, especially
for languages dissimilar to English
\citep{pires-etal-2019-multilingual}, and high variance
\citep{keung-etal-2020-dont}.

\paragraph{Semantic Parsing} Most previous work
\citep{lu-2014-semantic-hybrid-trees,neural-hybrid-trees-susanto2017,arch-for-neural-multisp-Susanto2017}
has focused on \emph{multilingual} semantic parsing, i.e., learning
from multiple natural languages in parallel, largely affirming the
benefit of ``high-resource'' multilingual data and multi-language
ensemble training \citep{multilingual-sp-hierch-tree-Jie2014}. 
\citet{shao-etal-2020-multi} further improved cross-lingual similarity
with adversarial language identification across such ensembled training data.
Code-switching in multilingual parsing has also been explored through mixed-language
training datasets
\citep{multilingsp-and-code-switching-Duong2017,einolghozati-etal-2021-el}.
To adapt a parser to new languages, machine translation has been
used as a reasonable proxy for in-language data
\citep{sherborne-etal-2020-bootstrapping,
  moradshahi-etal-2020-localizing}. However, machine translation, in
either direction can introduce limiting artifacts
\citep{artetxe-etal-2020-translation} with poor generalization due to
how ``translationese'' training data diverges from gold test utterances
\citep{riley-etal-2020-translationese}. 

Zero-shot parsing has primarily focused on `cross-domain' challenges
to improve generalization across varying query structures and lexicons
\citep{decouple-zero-shot-Herzig2018, givoli-reichart-2019-zero} or
different databases \citep{zhong-etal-2020-grounded-gazp,
  suhr-etal-2020-exploring,yu-etal-2018-spider}. The combination of
zero-shot parsing with cross-lingual modeling has also been examined
for the UCCA formalism \cite{hershcovich-etal-2019-semeval-task1} and
for task-oriented dialogue systems (see below).

\paragraph{Dialog Modeling} Cross-lingual transfer has been studied in
the context of goal-oriented dialog for the spoken language
understanding (SLU) tasks of intent classification and slot labeling
(i.e.,~parsing an utterance into a semantic frame identifying the
user's intent and its arguments). Recently released multilingual
datasets like MultiATIS++ \citep{xu-etal-2020-end-multiatis} and MTOP
\citep{li-etal-2021-mtop} have facilitated the study of zero-shot
transfer through the combination of pre-training, machine translation,
and word alignment (to project annotations between languages). Recent
work in this setting \citep{zhu-etal-2020-dont-parse-insert,
  li-etal-2021-mtop, krishnan2021multilingual-slu,
  nicosia-etal-2021-translate-fill-zero-shot-semparse} identifies a
penalty for cross-lingual transfer that neither pre-training nor
machine translation can fully overcome.



\section{Problem Formulation}
\label{sec:background}

The primary challenge for cross-lingual parsing is learning
parameters that can parse an utterance, $x$, from an unseen test
language to an accurate logical form (LF). 
Typically, a parser trained on language $l$, or multiple
languages $\lbrace l_1, \ldots, l_N\rbrace$, is only capable for
these languages and performs poorly outside this set. For a new
language, prior approaches require parallel datasets and models
\citep{
  multilingual-sp-hierch-tree-Jie2014,nlmaps-Haas2016,multilingsp-and-code-switching-Duong2017}.

In our work, zero-shot parsing refers to parsing utterances in new
languages \emph{without paired data during training}, For some
language, $l$, there exists no pairing of $x_l$ to a logical form,
$y$, except for English.\footnote{English is the ``source'' language
  for all our experiments. We refer to languages seen only at test
  time as ``new''.}  This setting also excludes ``silver-standard''
training pairs created using machine-translation.  As these models
have ultimately observed some form of utterance-logical form pairs for
each new language, we do not consider such approaches here and refer
to \citet{sherborne-etal-2020-bootstrapping} as an example of using MT
for this task.

It might be tempting to approach this problem as a case of fine-tuning
a pre-trained (English) decoder for LF generation.  Problematically,
the output target is expressed in a formally defined language
(e.g.,~SQL or $\lambda-$DCS) which models the semantics of questions
very differently to natural language (e.g.,~without presumption or
co-operation;
\citealt{kaplan-on-the-difference-between-nl-and-query}).  Formal
languages \citep{kamp1993discourse} additionally present artifacts
which render fine-tuning challenging such as unfamiliar syntax
(e.g.,~table aliases or explicit recursion) 
and long output sequences. In practice, we
observed fine-tuning leads to poor performance (e.g., \mbox{$<1\%$}~accuracy
on all languages), with the model insisting on hallucinating natural
language.  This is seemingly at odds with adjacent work in dialog
modeling, which has found pre-trained decoders to be beneficial
\citep{li-etal-2021-mtop}.  However, SLU requires learning a
lightweight label vocabulary compared to the $200+$ tokens required in
LFs. Additionally, SLU typically maintains output sequences of similar
size to natural language inputs (with tightly coupled syntactic
compositionality between the two), whereas the syntactic and
structural demands of LF generation are largely divorced from the
input utterance.

\begin{figure}[t!]
    \centering
    \includegraphics[width=\columnwidth]{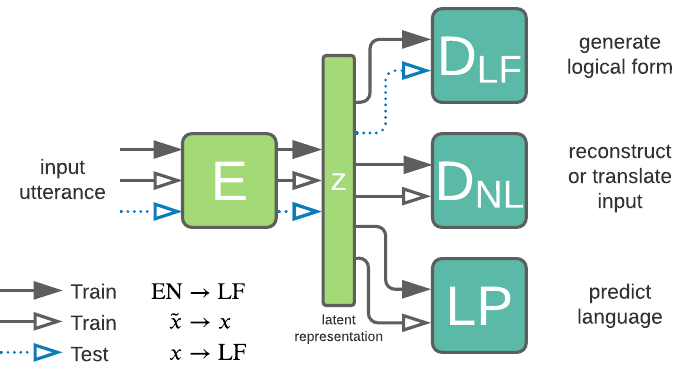}
    \caption{Our model, {\sc \bf ZX-Parse}, is a {\bf Z}ero-shot {\bf
        Cross}-lingual semantic {\bf Parse}r which augments an
      encoder-decoder model with auxiliary objectives.  The
      Encoder,~$E$, generates a representation,~$z$, which is input to
      the logical form decoder,~$D_{\rm LF}$, reconstruction
      decoder,~$D_{\rm NL}$, or language prediction
      classifier,~${\rm LP}$.  During training, English is input to
      all objectives and additional languages are incorporated using
      \emph{only} the additional objectives
      $\lbrace D_{\rm NL},~{\rm LP}\rbrace$. Logical forms are predicted
      using $D_{\rm LF}$ for all inputs at test time.}
    \label{fig:model}
\end{figure}

In our solution, the model is trained to parse from utterance-logical
forms pairs only in English. Other languages are incorporated using
auxiliary objectives and data detailed in Section~\ref{sec:method}. We
explore the hypothesis that an overlapping multi-lingual latent space
can be learned through auxiliary objectives in tandem with logical
form generation (see Figure~\ref{fig:model}). Our intuition is that
introducing these additional losses minimizes cross-lingual variance
in latent encoding space by optimizing for language-agnostic
representations with high similarity to the source language
(i.e.,~English). Our approach minimizes the cross-lingual transfer
penalty such that the zero-shot parser predicts logical forms from
test inputs regardless of utterance language.

By framing the cross-lingual parsing task as a latent
  representation alignment challenge, we explore a possible upper
bound of parsing accuracy without errors from external
dependencies. Section~\ref{sec:results} demonstrates that our zero-shot model,
using only English paired data and a small additional corpus, 
can generate accurate logical forms above translation baselines
to compete with fully supervised in-language training.



\section{Our Zero-shot Model:~{\sc ZX-Parse}}
\label{sec:method}

We adopt a multi-task sequence-to-sequence model
\citep{luong-2016-multitask-seq2seq} which combines logical form
generation with two auxiliary objectives. The first is a language
identification discriminator and the second is a reconstruction or translation decoder. 
An overview of our
semantic parser is given in Figure~\ref{fig:model}; we describe each
component below.

\paragraph{Generating Logical Forms}

Predicting logical forms is the primary objective for our model. Given
an utterance $x=\left(x_1,x_2,\ldots,x_T\right)$, we wish to generate
logical form $y=\left(y_1, y_2, \ldots, y_M\right)$ representing the
same meaning in a machine-executable language. We model this
transduction task using an encoder-decoder neural network
\citep{seq2seq-DBLP:journals/corr/SutskeverVL14} based upon the
Transformer architecture
\citep{transformers-noam-Vaswani2017AttentionIA}.

The sequence $x$ is encoded to a latent representation
$z=\left(z_1,z_2,\ldots,z_T\right)$ through Equation~\eqref{eq:latent}
using a stacked self-attention Transformer encoder,~$E$, with
weights~$\theta_{E}$.

\begin{align}
\centering
z &= E\left(x|\theta_{E}\right) \label{eq:latent} \\
p\left(y|x\right) &= \prod_{i=0}^{M}~p\left(y_{i}|y_{<i},
  x\right) \label{eq:output_seq_prob1} \\
  p\left(y_{i}|y_{<i},x\right) &= \operatorname{soft}\left(D_{\rm LF}\left(y_{<i}|z, \theta_{D_{\rm LF}}\right)\right) \label{eq:output_seq_prob2} \\
  \mathcal{L}_{\rm LF} &= -\sum_{\left(x,~y\right)\in\mathcal{S}_{\rm LF}}{\rm log}~p\left(y|x\right) \label{eq:lf_loss}
\end{align}
The conditional probability of the output sequence~$y$ is expressed in
Equation~\eqref{eq:output_seq_prob1} as each token~$y_i$ is
autoregressively generated based upon~$z$ and prior outputs, $y_{<i}$.
Equation~\eqref{eq:output_seq_prob2} models distribution~$p\left(y_{i}|y_{<i}, x\right)$ 
using a Transformer decoder for logical forms,~$D_{\rm LF}$, with associated weights
$\theta_{D_{LF}}$ where {$\operatorname{soft}$} is the softmax function.

We predict an output, $\hat{y}$, for semantic parsing
dataset $\mathcal{S}_{\rm LF}=\lbrace x^{n}, y^{n}\rbrace_{n=0}^{N}$,
through the encoder and logical form decoder, $\lbrace E,~D_{LF}
\rbrace$. Equation \eqref{eq:lf_loss} describes the loss 
objective minimizing the cross-entropy between $y$ and~$\hat{y}$.

\paragraph{Language Prediction}

Our first additional objective encourages language-agnostic
representations by reducing the discriminability of the source
language, $l$, from~$z$. Equation~\eqref{eq:dp_net} defines a {\bf
  L}anguage {\bf P}rediction (LP) network to predict $l$ from $z$
using a linear classifier over $L$~training languages:

\begin{equation}
{\rm LP}\left(x\right) = W_{i}x + b_{i} \label{eq:dp_net}
\end{equation}
where $W_{i} \in \mathbb{R}^{L\times|z|}$ and $b_{i} \in
\mathbb{R}^{L}$ are a weight and bias respectively. We
follow the best model from \citet{ahmad-etal-2019-cross}.
Equation \eqref{eq:dp_distribution} describes the conditional model
for the output distribution where a language label is predicted using
the time-average of the input encoding $z$ of length $T$: 

\begin{equation}
p\left(l|x\right) = \operatorname{soft}\left({\rm
    LP}\left(\frac{1}{T}\sum_{t} z_{t}
  \right)\right) \label{eq:dp_distribution} 
\end{equation}

Finally, Equation~\eqref{eq:dp_loss} describes the objective function
for the LP network:

\begin{equation}
\mathcal{L}_{\rm LP} = - \sum_{x} {\rm log}~p\left(l|x\right) \label{eq:dp_loss} 
\end{equation}
However, we \emph{reverse this gradient} in the
backward pass before the ${\rm LP}$ network, to encourage the
encoder to produce language invariant representations
\citep{gradreversal-Ganin:2016}. The ${\rm LP}$ network is optimized
to discriminate the source language from $z$, but the encoder is now
optimized adversarially \emph{against} this objective. Our intuition is 
that discouraging language discriminability in~$z$ encourages latent representation
similarity across languages, and therefore reduces the penalty for
cross-lingual transfer.

\paragraph{Generating Natural Language}

The final objective acts towards both regularization and cross-lingual
similarity. Motivated by \emph{domain-adaptive pre-training}
\citep{gururangan-etal-2020-dont-stop-pretraining}, we further adapt
the encoder towards question-style utterances from native speakers of
each test language lacking task-specific training data. We add an
additional Transformer decoder optimized to reconstruct a noisy input
from latent representation~$z$, in
Equation~\eqref{eq:latent}. Utterance, $x$, is input to the encoder,
$E$, and a separate decoder, $D_{\rm NL}$, then reconstructs~$x$
from~$z$. We follow the \emph{denoising} objective from
\citet{lewis-etal-2020-bart} and replace~$x$ with noised input
$\tilde{x}={\rm N}\left(x\right)$ with noising function~${\rm N}$.
The output probability of reconstruction is given in
Equation~\eqref{eq:denoise_prob_1} with each token predicted through
Equation~\eqref{eq:denoise_prob_2} using decoder, $D_{\rm NL}$, with
 weights~$\theta_{D_{\rm NL}}$:

\begin{align}
\centering
\hat{z} &= E\left(\tilde{x}|\theta_{E}\right) \label{eq:encoder_noised} \\
p\left(x|\tilde{x}\right) &= \prod_{i=0}^{T}~p\left(x_{i}|x_{<i}, \tilde{x}\right)  \label{eq:denoise_prob_1}  \\
\hspace*{-1.68ex}p\left(x_{i}|x_{<i},\tilde{x}\right) &\hspace{-.5ex}=\hspace{-.5ex} \operatorname{soft}\left(D_{\rm NL}\left(x_{<i}|\hat{z}, \theta_{D_{\rm NL}}\right)\right) \label{eq:denoise_prob_2} 
\end{align}
\newline \indent The auxiliary objectives are trained using both the
utterances from $\mathcal{S}_{\rm LF}$ and monolingual data,
$\mathcal{S}_{\rm NL}=\lbrace \lbrace x^{n} \rbrace_{n=0}^{N}
\rbrace_{l=0}^{L}$, in $L$ languages  (see Section \ref{sec:data}). 
 Submodel, $\lbrace E,~D_{\rm  NL} \rbrace$, predicts the 
reconstruction of~$x$ from~$\tilde{x}$ with the following 
objective:

\begin{equation}
\mathcal{L}_{\rm NL} = -\sum_{x}{\rm log}~p\left(x|\tilde{x}\right) \label{eq:nl_loss}
\end{equation}

In the form described above, this objective requires only unlabeled, monolingual
utterances in each target language. However, we can also augment it with a translation component to exploit natural language 
bi-text between the new language and English (e.g.,~$\mathcal{S}_{\rm NL}=\lbrace \lbrace
x_{\rm EN}^{n},~x_{l}^{n} \rbrace_{n=0}^{N} \rbrace_{l=0}^{L}$) to further
promote cross-lingual similarity.  According to some
sampling factor $\tau$, we randomly choose whether to
\emph{reconstruct} an utterance (as above) or \emph{translate} to the
parallel English utterance (i.e.,~replace~$x$ in
Equation~\eqref{eq:nl_loss} with $x_{\rm EN}$).

\paragraph{Combined Model}

The combined model uses a single encoder, $E$, and the three objective
decoders $\lbrace D_{\rm LF}, D_{\rm NL}, {\rm LP}\rbrace$ (see
Figure~\ref{fig:model}).  During training, an English query is encoded
and input to all three objectives to express output loss as
$\mathcal{L}_{\rm LF} + \mathcal{L}_{\rm NL} + \mathcal{L}_{\rm
  LP}$. For new languages without $\left(x, y\right)$ pairs, the
utterance is encoded and input only to the auxiliary objectives for a
combined loss as $\mathcal{L}_{\rm NL} + \mathcal{L}_{\rm LP}$. During
inference, an utterance is encoded and \emph{always} input to $D_{\rm
  LF}$ to predict a logical form, $\hat{y}$, regardless of test
language,~$l$. During the backward pass, each output loss
back-propagates the gradient signal from the respective objective
function. For the encoder, these signals are combined as:

\begin{align}
\hspace*{-1.5ex}\frac{\partial \mathcal{L}}{\partial \theta_E} = \frac{\partial \mathcal{L}_{\rm LF}}{\partial \theta_E}& - \lambda\alpha_{\rm LP}\frac{\partial \mathcal{L}_{\rm LP}}{\partial \theta_E} + \alpha_{\rm NL}\frac{\partial \mathcal{L}_{\rm NL}}{\partial \theta_E} \label{eq:encoder_loss} \\
\lambda &= \frac{2}{1+e^{-\gamma p}} - 1 \label{eq:lambda_schedule} \
\end{align}\\
where $\alpha_{\lbrace\rm LP,~NL\rbrace}$ are
loss weightings for auxiliary objectives and $\lambda$ is the reversed
gradient scheduling parameter from \citet{gradreversal-Ganin:2016}. 
The $\lambda$ value increments with training progress~$p$, 
scaled by $\gamma$, according to Equation~\eqref{eq:lambda_schedule},~
to limit the impact of noisy predictions during early training.

We expect that the parser will adapt and recognize an encoding
from an unfamiliar language through our joint training process, and
successfully connect new language representations to the logical-form
decoder at test time. This sequence-to-sequence approach is highly
flexible and may be useful for zero-shot approaches to additional
generation tasks (e.g., paraphrasing).



\section{Experimental Setup}
\label{sec:data}

\paragraph{Semantic Parsing Datasets} Our experiments examine whether
our zero-shot approach generalizes across languages and domains. We
evaluate performance on a new version of the \textbf{ATIS} dataset of
travel queries
\citep{hemphill-etal-1990-atis,atis-Dahl:1994:ESA:1075812.1075823}. We
align existing English utterances and SQL logical forms from
\citet{iyer-etal-2017-learning-user-feedback} to the multi-lingual
utterances from the MultiATIS++ dataset for spoken language
understanding \citep{xu-etal-2020-end-multiatis}. This alignment adds
executable SQL queries to utterances in Chinese (ZH), German (DE),
French (FR), Spanish (ES), and Portuguese (PT). We use the same
4,473/493/448 dataset split for training/validation/test as
\citet{kwiatkowski-etal-2011-lexical}.  We also add to the test set
Hindi (HI) and Turkish (TR) utterances from
\citet{Upadhyay2018-multiatis}.\footnote{Misalignment between ATIS
  versions result in the test sets containing 442 and 381 utterances
  for HI and TR respectively.} We can now predict SQL from the ATIS
test questions in eight natural languages. The MultiATIS++ Japanese
set was excluded as the utterance alignment to this language was not
recoverable.

We also examine \textbf{Overnight} \citep{Overnight-Wang15}, an
eight-domain dataset covering \emph{Basketball}, \emph{Blocks},
\emph{Calendar}, \emph{Housing}, \emph{Publications}, \emph{Recipes},
\emph{Restaurants}, and \emph{Social Network} domains. Overnight
comprises 13,682 English utterances paired with $\lambda-$DCS logical
forms, executable in SEMPRE \citep{SEMPRE-berant2013freebase}, split
into 8,754/2,188/2,740 for training/validation/test respectively. This
training data exists only in EN and we use the
ZH and DE test data from
\citet{sherborne-etal-2020-bootstrapping} for multilingual
evaluation. Given the varying linguistic phenomena across domains 
(e.g. relative spatial reasoning in \emph{Blocks} or temporal arithmetic in \emph{Calendar}), this dataset presents a harder challenge for cross-lingual transfer.

We measure performance with \emph{denotation accuracy} as all inferred
logical forms are executable in some knowledge base. This metric
compares the retrieved denotation from the prediction, $\hat{y}$, to
that from executing the gold-standard logical form.  Dataset sizes are
outlined in Appendix~\ref{app:more_setup}.

\paragraph{Natural Language Data}

For the reconstruction objective, we used the MKQA corpus \citep{mkqa}, a
multi-lingual translation of 10,000 samples from NaturalQuestions
\citep{kwiatkowski-etal-2019-natural}. This is suitable for our
auxiliary objective as the utterances are native-speaker question
surface forms, matching our test set while varying in subject. 
MKQA is also balanced across new languages to limit overexposure 
bias to one new language.  For bi-text, we use the original English and the
professionally translated question as a pair.

We also report experiments using a sample of crawled data from
ParaCrawl 7.1 \citep{banon-etal-2020-paracrawl}. The sample comprises
10,000 web scraped sentences paired with equivalent English to form
bi-text. Note that these samples are mostly declarative sentences and
as such do not match the surface form of our test inputs
(i.e.,~questions) and are also not parallel between sampled languages.
We contrast this to MKQA to examine how the \emph{style} of natural
language data influences performance.

For ATIS experiments, we use 60,000 utterances from each source in
languages with training data (EN, FR, PT, ES, DE, ZH). 
For Overnight, we use 30,000 utterances in EN, DE, and ZH.



\paragraph{Model Configuration}

The implementation of {\sc ZX-Parse} (see Section~\ref{sec:method})
largely follows parameter settings from
\citet{liu-etal-2020-multilingual-denoising} for Transformer encoder
and decoder layers (see Appendix~\ref{app:more_setup} for details on model
configuration). {\sc ZX-Parse} requires an encoder model to generate
multi-lingual latent representations for all objectives. Our
main results use only the encoder component of \emph{mBART50}
\citep{tang2020multilingua-mbart50} 
and we present experiments using other
pre-trained models in Appendix~\ref{app:more_results}.
We use all pre-trained encoder layers and 
append one additional learnable layer. All decoders are randomly
initialized six-layer stacks. Early experiments found this 
approach superior to any pre-trained decoder initialization.

The language predictor follows from \citet{ahmad-etal-2019-cross} 
as a single linear classification layer mapping from 1,024 inputs 
to $L$ output languages. Earlier findings 
supported that if the ${\rm LP}$ network is larger, then the reversed
gradient signal is too strong and therefore less useful as the ${\rm LP}$ 
network can memorize the language.

\paragraph{Comparison Models}
We primarily compare to a ``Translate-Test'' back-translation baseline
wherein the new language test set is translated to English using
Google Translate \citep{gtranslate} and input to a reference
sequence-to-sequence model trained on English. We also
compare to ``Translate-Train'', where we use MT from English to generate a
proxy dataset in each new language (e.g.,~French, Portuguese, Spanish,
German, Chinese, Hindi and Turkish) to train a monolingual parser.
We consider improving upon these ``minimum effort'' baselines 
as a lower bound for justifying our approach.

Additionally, we compare to an upper-bound monolingual model trained on
professional translations of the new languages.  We report results
on MultiATIS++ for FR, PT, ES, DE, and ZH (professional translations
are not available for Overnight training data). This is the
``maximum effort'' strategy that we desire to avoid. Parameters
for these reference systems match those outlined above
e.g.,~\emph{mBART50} encoder to logical form decoder.



\begin{table*}[ht!]
\centering
\begin{tabular}{@{}lcccccccc||ccc@{}}
\toprule
 & \multicolumn{8}{c||}{ATIS} & \multicolumn{3}{c}{Overnight} \\ 
 & EN & FR & PT & ES & DE & ZH & HI & TR & EN & DE & ZH \\ \midrule
Monolingual Training & 77.2 & 67.8 & 66.1 & 64.1 & 66.6 & 64.9 & --- & --- & 80.5 & --- & --- \\
Translate-Train & --- & 55.9 & 56.1 & 57.1 & 60.1 & 56.1 & \textbf{56.3} & 45.4 & --- & 62.2 & 59.4 \\
Translate-Test & --- & 58.2 & 57.3 & 57.9 & 56.9 & 51.4 & 52.6 & \textbf{52.7} & --- & 60.1 & 48.1 \\
{\sc ZX-Parse} & \textbf{76.9} & \textbf{70.2} & \textbf{63.4} & \textbf{59.7} & \textbf{69.3} & \textbf{60.2} &  54.9 & 48.3 & \textbf{81.9} & \textbf{66.2} & \textbf{60.0} \\ \bottomrule
\end{tabular}%

\caption{
  Denotation accuracy for ATIS
  \citep{atis-Dahl:1994:ESA:1075812.1075823} and Overnight (eight-domain
  average; \citealp{Overnight-Wang15})  for  supervised
  monolingual upper-bound, Translate-Test,  and our best
  {\sc ZX-Parse} model. Results for English (EN), French
  (FR), Portuguese (PT), Spanish (ES), German (DE), Chinese (ZH), Hindi
  (HI) and Turkish (TR) ranked by similarity to English
  \citep{ahmad-etal-2019-difficulties}. Best results compared to baselines are bolded. 
\label{tab:results_q1}
}
\end{table*}

\begin{table*}[ht!]
\centering
\begin{tabular}{@{}lcccccccc||ccc@{}}
\toprule
 & \multicolumn{8}{c||}{ATIS} & \multicolumn{3}{c}{Overnight} \\ 
{\sc ZX-Parse} & EN & FR & PT & ES & DE & ZH & HI & TR & EN & DE & ZH \\ \midrule
(a)~$D_{\rm LF}$ only & 77.2 & 61.3 & 42.5 & 46.5 & 50.2 & 38.5 & 40.4 & {37.3} & 80.5 & 58.4 & 48.0 \\
(b)~$D_{\rm LF}+D_{\rm NL}$ & \textbf{77.7} & 62.7 & 54.9 & 58.2 & 61.1 & 51.2 & 49.5 & {44.7} & 81.3 & 62.7 & 49.5 \\
(c)~$D_{\rm LF}+{\rm LP}$ & 76.3 & 57.2 & 53.7 & 51.8 & 58.6 & 44.1 & 39.8 & 38.8 & 80.6 & 60.6 & 49.4 \\
(d)~$D_{\rm LF}+{\rm LP}+D_{\rm NL}$ &  76.9 & \textbf{70.2} & \textbf{63.4} & \textbf{59.7} & \textbf{69.3} & \textbf{60.2} &  54.9 & 48.3 & \textbf{81.9} & \textbf{66.2} & \textbf{60.0}  \\ \bottomrule
\end{tabular}%
\caption{Denotation accuracy for ATIS and Overnight (eight-domain
  average) comparing ablations of {\sc ZX-Parse}:~(a) no
  auxiliary objectives,~(b) logical form (LF) generation and
  reconstruction,~(c) LF generation and language prediction,~(d) all objectives. Best results compared to baselines are bolded.\label{tab:results_q2} }
\end{table*}

\section{Results}
\label{sec:results}

Our results are outlined to answer four core questions, with
additional ablations in Appendix~\ref{app:more_results}.
Our findings support the hypothesis that we can minimize the
cross-lingual transfer penalty by improving latent alignment with
auxiliary objectives. We also examine the latent space directly 
and find {\sc ZX-Parse} learns more similar representations 
between languages. Our parser achieves state-of-the-art 
zero-shot results for all non-English languages in the 
MultiATIS++ and Overnight benchmarks.

\paragraph{Better than Translation?}

We compare between {\sc ZX-Parse} and the upper- and lower-bounds in
Table \ref{tab:results_q1}. Our multi-task approach significantly
improves upon ``Translate-Test'' for all languages included within the
auxiliary objectives ($p<0.01$). For ATIS, we find that
``Translate-Train'' performs below ``Translate-Test'' for languages
similar to English (FR, ES, PT) but worse for more distant languages
(DE, ZH).  {\sc ZX-Parse} performance improves on
``Translate-Train'' for all languages included in reconstruction (EN,
FR, PT, ES, DE, ZH), however, the general cross-lingual
improvement insufficiently extends to additional languages (HI, TR) to
perform above baselines.

Within {\sc ZX-Parse}, French and German demonstrate the
best zero-shot accuracy --- performing $+2.4\%$ and $+2.7\%$ above the
monolingual upper bound for ATIS. We do not observe similar
improvement for Portuguese or Spanish despite their similarity to
English.  This may be a result of German and French dominating the
pre-training corpora compared to other new languages.
\citep[their Table 6]{tang2020multilingua-mbart50}.

Our model demonstrates similar significant improvement for Overnight
($p<0.01$), however, we find lesser gain compared to ATIS. This may be
a consequence of the compounded challenge of evaluating eight varied
domains of complex linguistic constructs. Here, we find that 
``Translate-Train'' is a stronger approach than ``Translate-Test'',
which may be a consequence of machine-translation direction.
Our best approach on German still improves above ``Translate-Train'' 
($+4.0\%$), however, we find performance on Chinese to be only 
marginally improved by comparison ($+0.6\%$). We also observe some contrast
in {\sc ZX-Parse} performance related to orthographic similarity to
English. Parsing accuracy on Overnight in German is $+6.2\%$ above
Chinese, with a similar $+9.1\%$ gap between these same languages for ATIS.

\paragraph{Which Objective Matters?}

Ablations to the model are shown in Table~\ref{tab:results_q2},
identifying the contributions of different objectives. Model (a) shows
that without auxiliary objectives, performance in new languages is
generally below Translate-Test. This is unsurprising, as this
approach uses only pre-trained cross-lingual information without
additional effort to improve similarity. Such efforts are incorporated
in Model (b) using the additional reconstruction decoder. 
Even without the LP loss, domain targeted adaptation (with translation)
improves cross-lingual parsing by an average across new languages 
of $+9.3\%$ for ATIS  and $+2.9\%$ for Overnight. 
Notably, we identified an optimal ratio of translation to 
reconstruction of 50\% (i.e.,~$\tau=0.5$).  This
suggests that both monolingual utterances (for domain-adaptive tuning) and
bi-text (for translation) contribute to the utility of our method 
beyond reliance on one technique.

Evaluating the {$\rm LP$} objective within Model (c) and (d), we find
the reversed gradient successfully reduces language discriminability.
For Model (d), language prediction accuracy during training peaks at 93\% after 2\% progress
and subsequently decreases to <8\% beyond 10\% of training. 
Language prediction accuracy for the test set is 7.2\%. We observe
a similar trend for Model (c). Comparing individual objectives, we
find the addition of the language predictor alone less helpful
than the reconstruction decoder. Comparing Model~(a) and~(c), we observe
a smaller average improvement on new languages of $+4.3\%$ for ATIS and
 $+1.8\%$ for Overnight. This suggests adaptation towards specific surface
 form patterns can be more effective here than modeling languages as discrete labels.

 Considering the combination of objectives in Model (d), we identify
 cumulative benefit to parsing with both objectives. Compared to Model
 (a), the full model improves by an average of $+16.3\%$ for ATIS and
 $+9.9\%$ for Overnight across new languages. Our findings support our
 claim that latent cross-lingual similarity can be improved using
 auxiliary objectives and we specifically identify that a combination
 of approaches yields superior parsing.  We suggest that this
 combination benefits from constructive interference, as the language
 prediction loss promotes invariance in tandem with multi-lingual
 generation tasks adapting the encoder to improve modeling the surface
 form (e.g.,~questions from native speakers) of the new language test
 data.

\begin{table*}[th!]
\centering
\begin{tabular}{@{}lcccccccc||ccc@{}}
\toprule
 & \multicolumn{8}{c||}{ATIS} & \multicolumn{3}{c}{Overnight} \\ 
Baselines & EN & FR & PT & ES & DE & ZH & HI & TR & EN & DE & ZH \\ \midrule
Translate-Train & --- & 55.9 & 56.1 & 57.1 & 60.1 & 56.1 & \textbf{56.3} & 45.4 & --- & 62.2 & 59.4 \\
Translate-Test & --- & 58.2 & 57.3 & 57.9 & 56.9 & 51.4 & 52.6 & \textbf{52.7} & --- & 60.1 & 48.1 \\ \midrule
\multicolumn{12}{l}{{\sc ZX-Parse}} \\ \midrule
MKQA $\tau=0.0$ & \multicolumn{1}{c}{76.3} & 67.1 & 60.5 & 58.2 & 68.3 & 59.2 & 54.1 & 47.1 & 81.3 & 64.3 & 52.7 \\
MKQA $\tau=0.5$ & \textbf{76.9} & \textbf{70.2} & \textbf{63.4} & \textbf{59.7} & \textbf{69.3} & \textbf{60.2} &  54.9 & 48.3 & \textbf{81.9} & \textbf{66.2} & \textbf{60.0}  \\ \midrule
ParaCrawl $\tau=0.0$ & \multicolumn{1}{c}{72.7} & 63.4 & 58.0 & 54.1 & 62.0 & 50.9 & 46.9 & 39.9 & 78.4 & 62.4 & 51.1 \\
ParaCrawl $\tau=0.5$ & 76.5 & 64.6 & 60.3 & 59.2 & 63.1 & 52.6 & 47.8
& 45.8 & 81.2 & 63.2 & 52.9 \\\bottomrule
\end{tabular}%

\caption{
Denotation accuracy for ATIS and Overnight (eight-domain average)
comparing between data sources: MKQA (questions) or sampled web data
from ParaCrawl. We additionally contrast between modeling corpora as
monolingual text ($\tau=0$) or partially as bi-text
($\tau=0.5$). Best results compared to baselines are bolded.  
\label{tab:results_q3}
}
\end{table*}

\begin{figure}[t]
    \centering
    \includegraphics[width=\columnwidth]{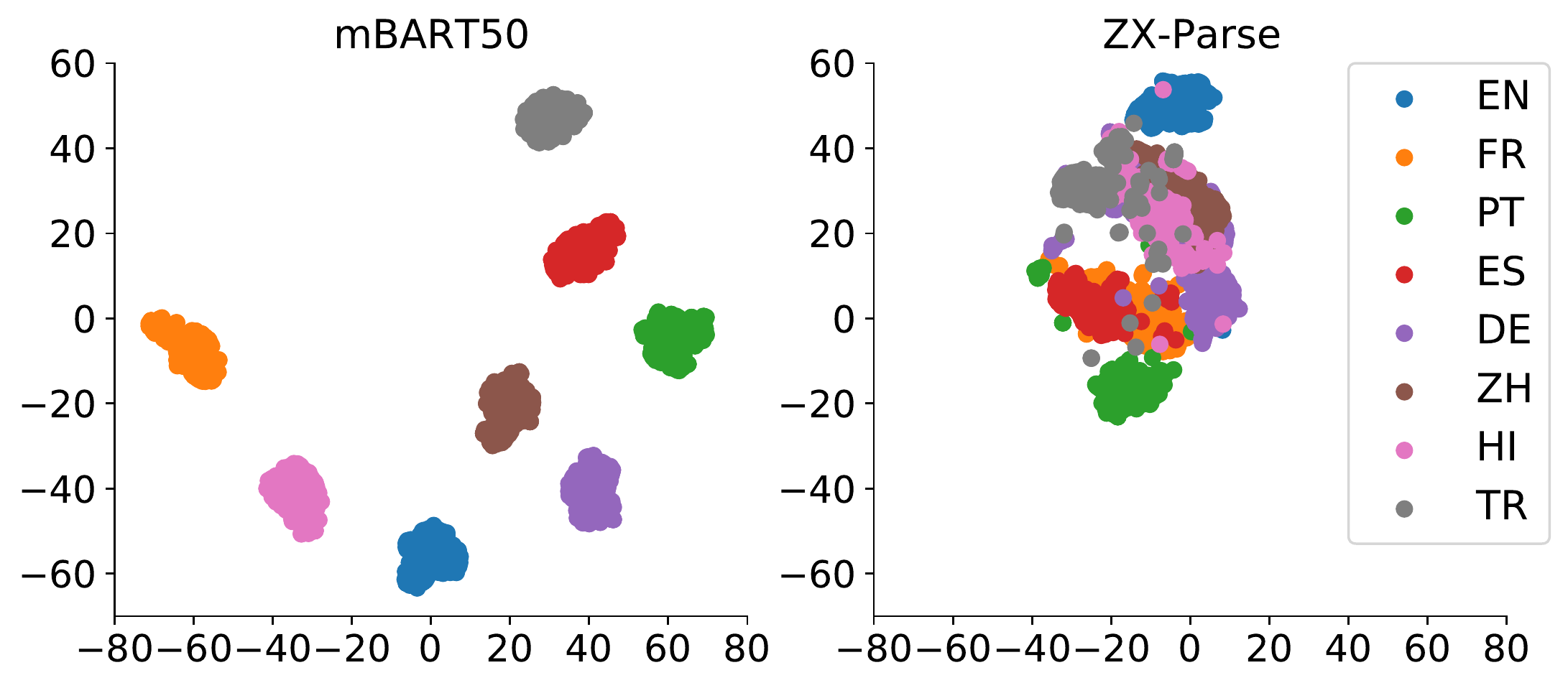}
    \caption{t-SNE comparison using mBART50 and {\sc ZX-Parse}
      encoders (MultiATIS++ test set). Our approach improves the
      latent alignment across languages.}
    \label{fig:tsne}
    \vspace{-1.5em} 
\end{figure}

Additional objectives also improve parsing for Hindi and Turkish
despite neither being included within auxiliary training data
(see HI and TR columns in Table~\ref{tab:results_q3}). By adapting our
latent representation to encourage similarity, we improve
parsing accuracy for two typologically diverse languages without
explicit guidance. To further examine this, we visualize the
MultiATIS++ test set in Figure~\ref{fig:tsne} and observe \emph{less
  discriminable} encodings from {\sc ZX-Parse} compared to
\emph{mBART50}. Quantitatively, we find the average cosine distance
between the sentence-mean of parallel utterances reduces from~0.58
to~0.47. Similarly, the average token-level symmetric Hausdorff
distance \cite{7053955_Hausdorff} between languages reduces from~0.72 to~0.41.  
This further supports that we learn more similar representations and our method has
wider utility beyond explicitly targeted languages.

\paragraph*{Does Language Style Matter?}

In Table~\ref{tab:results_q3} we examine whether our auxiliary
objectives are influenced by the style of natural language corpora for
reconstruction.  We find the use of questions positively improves
performance compared to crawled sentences. Using questions either as
monolingual utterances (i.e.,~no translation in~$D_{\rm NL}$) or with
as a bi-text sample (i.e.,~reconstruction and translation in $D_{\rm
  NL}$) improves above the Translate-Test baseline.  We observe modest
improvements with ParaCrawl, especially when introducing bi-text
into~$D_{\rm NL}$, but this is less consistent across
languages. Overall, our results suggest that {\sc ZX-Parse} is robust
even when question-style data is unavailable but can be particularly
effective when adapting towards both new languages \emph{and} domains.
We also examined the influence of language family on performance (see
Appendix~\ref{app:more_results}) and found that best performance
utilizes a linguistically varied ensemble of languages. Omitting
either Romance (ES/FR/PT) or Sino-Tibetan (ZH) languages in
reconstruction negatively impacts performance.

\paragraph{Where Does Improvement Come from?}

Comparing to Translate-Test, on ATIS, our best model generates 32\%
fewer ill-formed SQL requests and 24\% fewer extraneous queries
accessing unrelated tables in the database. Translation can fail when
entities are mishandled and our model generates 36\% fewer queries
with erroneous named entities. For Overnight, gains are strongly
related to improved numeracy in the model. Between our full model and
simplest approach (Model (a) in Table~\ref{tab:results_q2}), we find
more well-formed logical forms account for the largest improvement
(32.5\% fewer ill-formed SQL queries for ATIS and 35.2\% fewer
ill-formed $\lambda$-DCS queries for Overnight).  This supports our
notion in Figure~\ref{fig:problem} that better latent alignment can
minimize cross-lingual penalty.  However, improved structure
prediction is insufficient to solve this task on its own; 58.7\% of
remaining errors in the best model are
due to mishandled entities with the highest entity errors for Chinese
(60.2\%) and lowest for French (36.7\%). This suggests that aligning
entities across languages might be necessary for further improvement.



\section{Conclusion}

We presented a multi-task model for zero-shot cross-lingual semantic
parsing which combines logical form generation with auxiliary
objectives that require only modest natural language corpora for
localization. Through aligning latent representations, {\sc ZX-Parse}
minimizes the error from cross-lingual transfer and improves accuracy
across languages unseen during training. 

Although we focused exclusively on executable semantic parsing, our
approach is general and potentially relevant for linguistically
motivated frameworks such as Abstract Meaning Representation
\citep{banarescu-EtAl:2013:LAW7-ID,damonte2018cross} or Discourse
Representation Theory \citep{kamp1993discourse,evang2016cross}.  In
the future, we will investigate a few-shot scenario and study sample
efficient cross-lingual transfer by explicitly promoting
generalization using techniques such as meta-learning
\citep{pmlr-v70-finn17a}.


\section*{Ethics Statement}
 
A key limitation of our work is the limited coverage of eight higher-resource languages. As such, we are unable to test our approach in a \textbf{genuinely} low-resource scenario. We must also consider the risk of over-generalization to dominant dialects within each language as we lack an evaluation of additional dialects (e.g. our English dataset is representative of American English but not Indian English). We hope that such issues can be addressed with additional data collection. 

Our training requirements are detailed in Appendix \ref{app:more_setup}. We hope our work contributes to further usage and development of singular multilingual models as opposed to learning $N$ monolingual models for $N$ languages. 

\section*{Acknowledgements}
We thank the anonymous reviewers for their feedback and Bailin 
Wang, Kate McCurdy, and Rui Zhang for insightful discussion.
The authors gratefully acknowledge the support of 
the UK Engineering and Physical Sciences Research Council 
(grant EP/L016427/1; Sherborne) and the European Research 
Council (award number 681760; Lapata).

\bibliography{thesis}
\bibliographystyle{acl_natbib}

\appendix

\section{Experimental Setup}
\label{app:more_setup}

\begin{table}[th]
\centering
\begin{tabular}{@{}ccccc@{}}
\toprule
 & Train & Validation & Test & Total \\ \midrule
\multicolumn{5}{c}{ATIS} \\ \midrule
EN & 4,473 & 497 & 448 & 5,418 \\
FR & 4,473 & 497 & 448 & 5,418 \\
PT & 4,473 & 497 & 448 & 5,418 \\
ES & 4,473 & 497 & 448 & 5,418 \\
DE & 4,473 & 497 & 448 & 5,418 \\
ZH & 4,473 & 497 & 448 & 5,418 \\
HI & --- & --- & 442 & 442 \\
TR & --- & --- & 381 & 381 \\ \midrule
\multicolumn{5}{c}{Overnight} \\ \midrule
EN & 8,754 & 2,188 & 2,740 & 13,682 \\
DE & --- & 2,188 & 2,740 & 4,928 \\
ZH & --- & 2,188 & 2,740 & 4,928 \\ \bottomrule
\end{tabular}%

\caption{Semantic parsing dataset partitions per language for ATIS \citep{atis-Dahl:1994:ESA:1075812.1075823,Upadhyay2018-multiatis,xu-etal-2020-end-multiatis} and Overnight \cite{Overnight-Wang15,sherborne-etal-2020-bootstrapping}. Each example is an utterance paired with a logical form. 
}
\label{tab:dset_sizes}
\vspace{-1.5em}
\end{table}

\begin{table*}[!th]
\centering
\resizebox{\textwidth}{!}{%
\begin{tabular}{@{}lccccc@{}}
\toprule
Model & \# Layers & \# Parameters & \# Vocabulary & Tokenization & \# Languages \\ \midrule
mBART-large (encoder) & 12 & 408M & 250,027 & bBPE & 25 \\
XLM-R-large & 24 & 550M & 250,002 & bBPE & 100 \\
mBART50-large (encoder) & 12 & 408M & 250,054 & bBPE & 50 \\ \midrule
 & &  & 593
  (ATIS) &  & 8 (ATIS) \\
\raisebox{1.5ex}[0pt]{{\sc ZX-Parse}} &  \raisebox{1.5ex}[0pt]{6 (decoder)} & \raisebox{1.5ex}[0pt]{208M} &  226 (Overnight) & \raisebox{1.5ex}[0pt]{Whitespace}
& 3 (Overnight) \\ \bottomrule
\end{tabular}%
}
\caption{Pretrained model configurations and configuration for the
  trainable components of {\sc ZX-Parse} (e.g.,~the objectives). All
  models use a hidden dimension of 1,024, a feed-forward hidden
  projection of 4,096 and 16 heads per multi-head attention layer. For
  natural language, all models use byte-level BPE tokenization
  \citep{wang2019neural} and logical forms are tokenized using
  whitespace.} 
\label{tab:pt_models}
\vspace{-1.5em}
\end{table*}

\paragraph{Zero-shot Model Configuration}

The encoder,~$E$, decoders, $\lbrace D_{\rm LF}, D_{\rm NL} \rbrace$,
and embedding matrices all use a dimension size of 1,024 with the
self-attention projection of 4,096 and 16~heads per layer. Both
decoders are 6-layer stacks. Weights were initialized by sampling from
normal distribution $\mathcal{N} \left( 0, 0.02 \right)$.  The
language prediction network is a two-layer feed-forward network
projecting from~$z$ to 1,024 hidden units then to~$|L|$ for~$L$
languages. $L$~is six for experiments on ATIS and three for
experiments on Overnight.

Configurations for models used in this work are reported in Table
\ref{tab:pt_models} with similar details for the objective components
of {\sc ZX-Parse}. Initial experiments examined \emph{XLM-R-base},
which is 12 layers opposed to 24, however, performance was
significantly worse and, therefore, this model was not considered
further. Experiments reported in Section \ref{sec:results} all use
\emph{mBART50} as the pre-trained encoder as all other pre-trained
models performed significantly worse (see
Appendix~\ref{app:more_results}).  In all our experiments, we found
that a randomly initialized decoder was superior to using pre-trained
weights.

A complete outline of dataset partitions per language is shown in
Table \ref{tab:dset_sizes} for both datasets. \mbox{{\sc ZX-Parse}}
uses only English training and validation data and tests on all
additional languages. We did not use multi-lingual validation data as
recommended in \citet{keung-etal-2020-dont} as this approach did not
prove critically beneficial in early experiments and doing so would
explode the data requirements for a multi-lingual system. 

\paragraph{Experimental Setting}

The system was trained using the Adam optimizer
\citep{ADAMOPT-Kingma2014AdamAM} with a learning rate of
$1\times10^{-4}$, and a weight decay factor of 0.1. We use a ``Noam''
schedule for the learning rate
\citep{transformers-noam-Vaswani2017AttentionIA} with a warmup of
5,000 steps. For pre-trained components, we fine-tune XLM-R and mBART
encoders with learning rate $1\times10^{-5}$ but freeze the encoder
when using mBART50. Loss weighting values for $\alpha_{\lbrace\rm
  LP,~NL\rbrace}$ were empirically optimized to $\lbrace 0.33,
0.1\rbrace$ respectively from a range $\lbrace 0.5, 0.33, 0.1, 0.05, 0.01, 0.005, 0.001\rbrace$. 
Batches during training were size 50 and
homogeneously sampled from either $\mathcal{S}_{\rm LF}$ or
$\mathcal{S}_{\rm NL}$, with an epoch consuming one pass over
both. Models were trained for a maximum of 100 epochs with early
stopping. Model selection and hyper-parameters were tuned on the
$\mathcal{S}_{\rm LF}$ validation set in English e.g.,~validation only
evaluates performance on logical-form generation and not additional
objectives. Test predictions were generated using beam search with 5
hypotheses.

For the reconstruction noising function,
we use token masking to randomly replace~$u$ tokens in $x$ with ``{\tt
<mask>}'' where $u$ is sampled from $U\left(0, v\right)$. We found $v=3$ 
as the empirically optimal maximum tokens to mask in an input. Similarly,
we found $\gamma~=40$ optimal for the language prediction loss and $\tau=0.5$
as the optimal sampling factor for translation versus reconstruction. This
value of $\tau$ corresponds to using half the reconstruction data as mono-lingual
utterances and half as bi-text paired with English. 

\paragraph{Reproducibility}

All models were implemented using AllenNLP \citep{AllenNLP} and
PyTorch \citep{pytorch}, using pre-trained models from HuggingFace
\citep{Wolf2019HuggingFacesTS}. Each model is trained on 1 NVIDIA
RTX3090 GPU in a cluster configuration, with no model requiring over
24 hours to complete training. Hyper-parameters were chosen by training
a reference model for parsing English utterances and selecting the
system with minimum validation loss. Our optimization grid-search
explored: $\lbrace6,~9,~12\rbrace$ decoder layers; freezing or
unfreezing the pre-trained encoder; $\lbrace0,~1,~2\rbrace$ additional
encoder layers appended to the pre-trained encoder; learning rates of
$1\times10^{\lbrace-3,~-4,~-5\rbrace}$ and a weight decay factor of
$\lbrace0,0.1,0.01\rbrace$. Optimal parameters in these early tests
were carried through for all additional models.

Additionally, we optimized hyper-parameters for auxiliary objectives
through linear search with all other factors fixed. The upper limit, $v$, for the
number of tokens to mask during reconstruction, $U\left(0, v\right)$,
was optimized from integers 1-6. The MKQA dataset used for auxiliary
tasks contains shorter sentences than prior work using masking, such
as \citet{lewis-etal-2020-bart}, and we observed that high levels of
masking ultimately destroys the input sentence and handicaps the
overall task. $\tau$ was optimized between values
of 0.0 (e.g. ignore bi-text) to 1.0 (e.g. all data is used as bi-text)
in increments of 0.1. Finally, we optimize the $\gamma$ parameter within
Equation \ref{eq:dp_loss} between
$\lbrace0,~5,~10,~20,~40,~50,~100\rbrace$ on an approximately
logarithmic scale. The optimal value of $\gamma=40$ results in loss
$\mathcal{L}_{\rm LF}$ reaching 99\% of the maximum value at approximately
13.6\% of training progress.



\begin{table*}[t]
\centering
\begin{tabular}{@{}lcccccccc@{}}
\toprule
  & EN & FR & PT & ES & DE & ZH & HI & TR \\
\midrule
Monolingual Training & 77.2 & 67.8 & 66.1 & 64.1 & 66.6 & 64.9 & --- & --- \\
Translate-Train & --- & 55.9 & 56.1 & 57.1 & 60.1 & 56.1 & \textbf{56.3} & 45.4 \\
Translate-Test & --- & 58.2 & 57.3 & 57.9 & 56.9 & 51.4 & 52.6 & \textbf{52.7} \\
\midrule
\multicolumn{9}{l}{{\sc ZX-Parse} using mBART} \\ \midrule
\quad$D_{\rm LF}$ only & 74.6 & 35.4 & 18.3 & 55.6 & 35.9 & 10.8 & 10.7 & 21.4 \\
\quad$D_{\rm LF} + D_{\rm NL}$ & 77.7 & 27.9 & 17.6 & 54.5 & 34.5 & 10.3 & 12.6 & 21.1 \\
\quad$D_{\rm LF} + {\rm LP}$ & 75.3 & 32.3 & 15.3 & 49.3 & 32.3 & 7.4 & 10.7 & 19.2 \\
\quad$D_{\rm LF} + {\rm LP}+D_{\rm NL}$ & 77.0 & 39.4 & 24.6 & 56.3 & 37.6 & 12.9 & 11.0 & 32.6 \\ \toprule
\multicolumn{9}{l}{{\sc ZX-Parse} using XLM-R} \\ \midrule
\quad$D_{\rm LF}$ only & 76.5 & 44.6 & 47.2 & 57.1 & 41.8 & 16.2 & 11.2 & 14.5 \\
\quad$D_{\rm LF} + D_{\rm NL}$ & {\bf 78.6} & 36.9 & 43.4 & 57.0 & 46.0 & 11.7 & 11.2 & 12.6 \\
\quad$D_{\rm LF} + {\rm LP}$ & 74.9 & 30.8 & 32.6 & 56.6 & 30.8 & 11.5 & 11.4 & 21.4 \\
\quad$D_{\rm LF} + {\rm LP}+D_{\rm NL}$ & 78.2 & 48.1 & 46.0 & {\bf
  60.8} & 55.4 & 18.3 & 18.3 & 35.8 \\ \toprule
\multicolumn{9}{l}{{\sc ZX-Parse} using mBART50} \\ \midrule
\quad$D_{\rm LF}$ only & 77.2 & 61.3 & 42.5 & 46.5 & 50.2 & 38.5 & 40.4 & 37.3 \\
\quad$D_{\rm LF} + D_{\rm NL}$ & 77.7 & 62.7 & 54.9 & 58.2 & 61.1 & 51.2 & 49.5 & 44.7 \\
\quad$D_{\rm LF} + {\rm LP}$ & 76.3 & 57.2 & 53.7 & 51.8 & 58.6 & 44.1 & 39.8 & 38.8 \\
\quad$D_{\rm LF} + {\rm LP}+D_{\rm NL}$ & 76.9 & \textbf{70.2} & \textbf{63.4} & 59.7 & \textbf{69.3} & \textbf{60.2} & 54.9 & 48.3 \\
\bottomrule
\end{tabular}
\caption{Complete denotation accuracy results for ATIS across all languages: English (EN), French (FR), Portuguese (PT), 
  Spanish (ES), German (DE), Chinese (ZH), Hindi (HI) and Turkish (TR). Models shown use (i) no auxiliary objectives, (ii) logical 
  form generation and reconstruction, (iii) logical form generation and language prediction and (iv) finally all objectives.
}
\label{tab:all_atis}
\vspace{-1.5em}
\end{table*}

\begin{table*}[t]
\centering
\begin{tabular}{@{}lccccccccc@{}}
\toprule
{\sc ZX-Parse} using mBART & Avg. & Ba. & Bl. & Ca. & Ho. & Pu. & Rec. & Res. & So. \\ \midrule
$D_{\rm LF}$ only & 81.0 & 89.0 & 64.7 & 81.5 & 77.8 & 77.6 & 88.0 & 86.1 & 83.6 \\
$D_{\rm LF} + D_{\rm NL}$ & 81.7 & 89.0 & 65.7 & {\bf 86.3} & 77.2 & 81.4 & 86.4 & 85.2 & 82.5 \\
$D_{\rm LF} + {\rm LP}$ & 79.8 & 85.9 & 65.9 & 82.1 & 74.1 & 77.0 & 88.0 & 83.3 & 81.9 \\
$D_{\rm LF} + D_{\rm NL}+ {\rm LP}$ & 80.5 & 86.7 & 65.7 & 84.5 & 75.1
& 82.6 & 85.8 & 83.3 & 80.3 \\ \toprule
{\sc ZX-Parse} using XLM-R & Avg. & Ba. & Bl. & Ca. & Ho. & Pu. & Rec. & Res. & So. \\ \midrule
 $D_{\rm LF}$ only & {\bf 82.3} & 89.3 & {\bf 67.7} & 85.1 & 75.7 & {\bf 85.7} & {\bf 89.8} & 81.0 & {\bf 84.4} \\
$D_{\rm LF} + D_{\rm NL}$ & 81.7 & 89.3 & 61.4 & 84.5 & 77.2 & 82.6 & 88.3 & {\bf 86.6} & 83.8 \\
$D_{\rm LF} + {\rm LP}$ & 76.7 & 77.0 & 61.7 & 74.4 & 75.7 & 74.5 & 86.1 & 85.2 & 79.4 \\
$D_{\rm LF} + D_{\rm NL}+ {\rm LP}$ & 82.2 & 87.7 & 64.9 & {\bf 86.3}
& 77.2 & 82.6 & 89.2 & 85.6 & 84.2 \\  \toprule
{\sc ZX-Parse} using mBART50 & Avg. & Ba. & Bl. & Ca. & Ho. & Pu. & Rec. & Res. & So. \\ \midrule
$D_{\rm LF}$ only & 80.5 & {\bf 90.0} &
66.7 & 82.7 & 76.7 & 75.8 & 87.7 & 83.3 & 80.9 \\ 
$D_{\rm LF} + D_{\rm NL}$ & 81.3 & 88.5 & 60.9 & 83.3 & {\bf 78.8} & 83.9 & 88.6 & 83.8 & 82.6 \\
$D_{\rm LF} + {\rm LP}$ & 80.6 & 90.0 & 63.4 & 78.6 & 76.2 & 81.4 & 84.7 & 86.4 & 83.8 \\
$D_{\rm LF} + D_{\rm NL}+ {\rm LP}$ & 81.9 & 87.7 & 65.4 & 84.5 & 77.8 & 81.4 & 88.0 & 87.0 & 83.1 \\

\bottomrule
\end{tabular}%
\caption{Denotation accuracy for the Overnight dataset \citep{Overnight-Wang15} from \textbf{English} utterances. Domains are \emph{Basketball}, 
\emph{Blocks}, \emph{Calendar}, \emph{Housing}, \emph{Publications}, \emph{Recipes}, \emph{Restaurants} and \emph{Social Network}.\label{tab:all_onight_en}} 
\vspace{-1.5em}
\end{table*}

\begin{table*}[t]
\centering
\begin{tabular}{@{}lrrrrrrrrr@{}}
\toprule
&  Avg. & Ba. & Bl. & Ca. & Ho. & Pu. & Rec. & Res. & So. \\ \midrule
Translate-Train & 62.2  & 73.5  & 45.4  & 68.0  & 49.3  & {\bf 64.0}  & 67.5  & 59.8  & 70.2 \\
Translate-Test & 60.1 & 75.7 & 50.9 & 61.0 & 55.6 & 50.4 & 69.9 & 46.3 & 71.4 \\ 
\midrule
\multicolumn{9}{l}{{\sc ZX-Parse} using mBART} \\ \midrule
$D_{\rm LF}$ only & 40.8 & 60.1 & 42.6 & 32.1 & 42.9 & 23.6 & 39.8 & 35.6 & 49.9 \\
$D_{\rm LF} + D_{\rm NL}$ & 39.0 & 55.8 & 41.9 & 39.3 & 35.4 & 20.5 & 38.9 & 32.9 & 47.2 \\
$D_{\rm LF} + {\rm LP}$ & 38.6 & 61.1 & 45.6 & 28.0 & 45.5 & 14.9 & 38.9 & 18.1 & 57.1 \\
$D_{\rm LF} + D_{\rm NL}+ {\rm LP}$ & 52.6 & 69.8 & 47.1 & 47.6 & 51.3 & 51.6 & 56.6 & 38.9 & 57.5 \\ \toprule
\multicolumn{9}{l}{{\sc ZX-Parse} using XLM-R} \\ \midrule
$D_{\rm LF}$ only & 38.6 & 45.0 & 43.4 & 21.4 & 45.5 & 32.3 & 37.7 & 39.8 & 44.1 \\
$D_{\rm LF} + D_{\rm NL}$ & 45.8 & 58.6 & 48.4 & 33.3 & 38.1 & 39.1 & 44.3 & 51.9 & 52.9 \\
$D_{\rm LF} + {\rm LP}$ & 41.1 & 64.2 & 41.4 & 26.2 & 31.2 & 38.5 & 39.2 & 44.4 & 43.8 \\
$D_{\rm LF} + D_{\rm NL}+ {\rm LP}$ & 49.0 & 68.3 & 48.6 & 48.2 & 42.3 & 46.6 & 48.2 & 34.7 & 55.4 \\ \toprule
\multicolumn{9}{l}{{\sc ZX-Parse} using mBART50} \\ \midrule
$D_{\rm LF}$ only & 58.4 & 70.3 & 51.1 & 61.9 & 54.0 & 49.7 & 65.4 & 42.1 & 73.1 \\
$D_{\rm LF} + D_{\rm NL}$ & 62.7 & 73.1 & 56.1 & 66.1 & 58.7 & 49.7 & {\bf 70.2} & 57.9 & 70.1 \\
$D_{\rm LF} + {\rm LP}$ & 60.6 & 76.5 & 57.9 & 68.5 & 52.9 & 52.8 & 36.6 & 66.6 & 73.2 \\
$D_{\rm LF} + D_{\rm NL}+ {\rm LP}$ & {\bf66.2} & {\bf79.8} & {\bf60.4} & {\bf72.6} & {\bf60.3} & 62.1 & 45.8 & {\bf74.4} & {\bf73.9} \\ \bottomrule

\end{tabular}
\caption{Denotation accuracy for the Overnight dataset \citep{Overnight-Wang15} using the \textbf{German} test set from \citet{sherborne-etal-2020-bootstrapping}. Domains are \emph{Basketball}, 
\emph{Blocks}, \emph{Calendar}, \emph{Housing}, \emph{Publications}, \emph{Recipes}, \emph{Restaurants} and \emph{Social Network}.\label{tab:all_onight_de}}
\vspace{-1.5em}
\end{table*}

\begin{table*}[t]
\centering
\begin{tabular}{@{}lccccccccc@{}}
\toprule
&  Avg. & Ba. & Bl. & Ca. & Ho. & Pu. & Rec. & Res. & So. \\ \midrule
Translate-Train & 59.4 & {\bf 75.4} & 46.5 & 50.5 & 57.8 & {\bf 56.7} & {\bf 62.1} & {\bf 60.1} & 66.1 \\ 
Translate-Test & 48.1 & 62.3 & 39.6 & 49.8 & 43.1 & 48.3 & 51.4 & 29.2 & 61.2 \\ 
\midrule
\multicolumn{9}{l}{{\sc ZX-Parse} using mBART} \\ \midrule
$D_{\rm LF}$ only & 19.4 & 11.5 & 23.6 & 23.8 & 33.9 & 6.8 & 21.1 & 12.5 & 22.4 \\
$D_{\rm LF} + D_{\rm NL}$ & 16.7 & 1.0 & 28.1 & 20.8 & 32.8 & 9.3 & 15.7 & 7.9 & 18.1 \\
$D_{\rm LF} + {\rm LP}$ & 17.0 & 15.9 & 26.8 & 7.7 & 28.0 & 2.5 & 24.4 & 5.6 & 25.5 \\
$D_{\rm LF} + D_{\rm NL}+ {\rm LP}$ & 36.1 & 26.1 & 30.6 & 42.9 & 49.2 & 28.6 & 44.9 & 23.1 & 43.2 \\
 \toprule
\multicolumn{9}{l}{{\sc ZX-Parse} using XLM-R} \\ \midrule
$D_{\rm LF}$ only & 17.6 & 6.1 & 24.8 & 20.2 & 21.7 & 14.9 & 22.6 & 20.4 & 10.2 \\
$D_{\rm LF} + D_{\rm NL}$ & 18.0 & 17.6 & 13.3 & 8.9 & 33.9 & 13.7 & 21.4 & 10.6 & 24.4 \\
$D_{\rm LF} + {\rm LP}$ & 18.5 & 20.5 & 4.3 & 13.1 & 39.7 & 5.0 & 19.3 & 13.4 & 32.5 \\
$D_{\rm LF} + D_{\rm NL}+ {\rm LP}$ & 22.7 & 18.4 & 30.3 & 15.5 & 37.0 & 13.0 & 19.6 & 7.9 & 39.5 \\
\toprule
\multicolumn{9}{l}{{\sc ZX-Parse} using mBART50} \\ \midrule
$D_{\rm LF}$ only & 48.0 & 53.7 & 49.6 & 53.0 & 50.8 & 36.0 & 52.1 & 23.1 & 65.3 \\
$D_{\rm LF} + D_{\rm NL}$ & 49.5 & 56.6 & 49.4 & 55.4 & 56.1 & 35.4 & 54.2 & 24.9 & 64.1 \\
$D_{\rm LF} + {\rm LP}$ & 49.4 & 55.5 & 54.9 & 73.8 & 53.4 & 19.3 & 21.3 & 52.7 & 63.9 \\
$D_{\rm LF} + D_{\rm NL}+ {\rm LP}$ & {\bf 60.0} & 59.4 & {\bf57.4} & {\bf74.4} & {\bf62.2} & 41.1 & 59.3 & 57.9 & {\bf68.4} \\ \bottomrule
\end{tabular}
\caption{Denotation accuracy for the Overnight dataset \citep{Overnight-Wang15} using the \textbf{Chinese} test set from \citet{sherborne-etal-2020-bootstrapping}. Domains are \emph{Basketball}, 
\emph{Blocks}, \emph{Calendar}, \emph{Housing}, \emph{Publications}, \emph{Recipes}, \emph{Restaurants} and \emph{Social Network}.\label{tab:all_onight_zh}}
\vspace{-1.5em}
\end{table*}

\begin{table*}[th]
\centering
\begin{tabular}{@{}lccccccccc@{}}
\toprule
 & Avg. & Ba. & Bl. & Ca. & Ho. & Pu. & Rec. & Res. & So. \\ \midrule
\multicolumn{10}{l}{EN} \\ \midrule
MKQA $\tau=0.0$ & 81.3 & {\bf89.3} & 63.9 & 82.7 & {\bf79.4} & {\bf82.6} & 83.3 & {\bf87.7} & 81.9 \\
MKQA $\tau=0.5$ & {\bf 81.9} & 87.7 & {\bf 65.4} & {\bf84.5} & 77.8 & 81.4 & {\bf88.0} & 87.0 & 83.1 \\ \midrule
ParaCrawl $\tau=0.0$ & 78.4 & 87.5 & 58.9 & 77.4 & 72.0 & 78.3 & 84.7 & 86.4 & 81.9 \\
ParaCrawl $\tau=0.5$ & 81.2 & 87.2 & 65.2 & {\bf84.5} & 74.6 & 80.7 & 86.6 & 87.0 & {\bf83.4} \\ \midrule
\multicolumn{10}{l}{DE} \\ \midrule
MKQA $\tau=0.0$ & 64.3 & 78.8 & 56.6 & 68.5 & 58.7 & 46.6 & {\bf70.8} & 59.3 & {\bf75.5} \\
MKQA $\tau=0.5$ & {\bf66.2} & {\bf79.8} & {\bf60.4} & {\bf72.6} & {\bf60.3} & {\bf62.1} & 45.8 & {\bf74.4} & 73.9 \\ \midrule
ParaCrawl $\tau=0.0$ & 62.4 & 76.3 & 53.7 & 64.4 & 57.6 & 51.3 & 50.1 & 72.3 & 73.6 \\
ParaCrawl $\tau=0.5$ & 63.2 & 78.0 & 55.9 & 69.0 & 57.1 & 56.5 & 44.0 & 70.2 & 74.4 \\ \midrule
\multicolumn{10}{l}{ZH} \\ \midrule
MKQA $\tau=0.0$ & 52.7 & 59.1 & 50.1 & 67.9 & 54.5 & {\bf41.6} & {\bf59.6} & 20.8 & 67.6 \\
MKQA $\tau=0.5$ & {\bf 60.0} & {\bf 59.4} & {\bf 57.4} & {\bf 74.4} & {\bf 62.2} & 41.1 & 59.3 & {\bf57.9} & 68.4 \\ \midrule
ParaCrawl $\tau=0.0$ & 51.1 & 56.3 & 55.5 & 64.8 & 54.5 & 26.6 & 27.0 & 53.8 & {\bf70.5} \\
ParaCrawl $\tau=0.5$ & 52.9 & 58.8 & 55.9 & 71.4 & 60.8 & 30.4 & 30.1 & 46.8 & 68.6 \\ \bottomrule
\end{tabular}
\caption{Denotation accuracy for the Overnight dataset \citep{Overnight-Wang15} compared across reconstruction data usage for English, German and Chinese. 
We compare between MKQA \citep{mkqa} and ParaCrawl \citep{banon-etal-2020-paracrawl} with additional contrast between using reconstruction data 
as monolingual utterances (e.g. $\tau=0.0$) or with some proportion as bi-text where the target sequence is replaced with the parallel English utterance (e.g. $\tau=0.5$).
Domains are \emph{Basketball}, 
\emph{Blocks}, \emph{Calendar}, \emph{Housing}, \emph{Publications}, \emph{Recipes}, \emph{Restaurants} and \emph{Social Network}. Best results for each language are bolded.} 
\label{tab:results_q3_overnight_full}
\end{table*}

\begin{table*}[th]
\centering
\begin{tabular}{@{}lcccccccc@{}}
\toprule
 & EN & FR & PT & ES & DE & ZH & HI & TR \\
 \midrule
Omit Romance Genus (FR, ES, PT)  & 74.1 & 59.0 & 58.0 & 52.6 & 64.4 & 56.3 & 45.1 & 39.7 \\
Omit Sino-Tibetan Family (ZH) & 74.1 & 65.1 & 58.3 & 55.8 & 65.1 & 43.2 & 42.7 & 38.7 \\ \bottomrule
\end{tabular}
\caption{Denotation accuracy for the ATIS dataset compared across reconstruction language ablations. 
We experiment with omitting the Romance genus (French, Spanish, Portuguese) and the Sino-Tibetan
family (ZH only). Language groupings are sourced from \citet{wals}.
} 
\label{tab:lang_family_ablation_atis}
\vspace{-1.5em}
\end{table*}

\begin{table*}[th]
\centering
\begin{tabular}{@{}llllllllll@{}}
\toprule
 & Avg. & Ba. & Bl. & Ca. & Ho. & Pu. & Rec. & Res. & So. \\
 \midrule
\multicolumn{10}{l}{Omit Sino-Tibetan Family (ZH)} \\ \midrule
EN & 80.7 & 87.7 & 63.9 & 81.5 & 77.8 & 82.0 & 85.6 & 84.6 & 82.2 \\ 
DE & 61.8 & 75.7 & 53.6 & 63.7 & 56.6 & 50.3 & 49.5 & 71.7 & 73.5 \\
ZH & 48.9 & 54.7 & 52.9 & 63.1 & 51.9 & 23.6 & 25.9 & 51.5 & 67.5 \\ \bottomrule
\end{tabular}
\caption{Denotation accuracy for the Overnight dataset compared across reconstruction language ablations. 
We report results for English, German and Chinese when omitting the Sino-Tibetan family (ZH only). 
Language groupings are sourced from \citet{wals} and domains are \emph{Basketball}, 
\emph{Blocks}, \emph{Calendar}, \emph{Housing}, \emph{Publications}, \emph{Recipes}, \emph{Restaurants} and \emph{Social Network}.
} 
\label{tab:lang_family_ablation_overnight}
\vspace{-1.5em}
\end{table*}

\section{Additional Results}
\label{app:more_results}

We extend the results in Section~\ref{sec:results} to include
additional ablations for all pre-trained
models. Table~\ref{tab:all_atis} details all results for ATIS across
eight test languages. Additionally, complete results across all
domains in Overnight are reported for English in Table
\ref{tab:all_onight_en}, German in Table \ref{tab:all_onight_de}, and
Chinese in Table~\ref{tab:all_onight_zh}. Table \ref{tab:results_q3},
comparing between reconstruction data sources, is expanded on for
Overnight in Table \ref{tab:results_q3_overnight_full}. Finally, we
present additional ablations to our model considering reconstruction
language families in Table~\ref{tab:lang_family_ablation_atis}
and~\ref{tab:lang_family_ablation_overnight}.

\paragraph*{Which Pre-trained Encoder?}

Our full results using three different pre-trained encoders
are outlined in Tables \ref{tab:all_atis}--\ref{tab:all_onight_zh}.
Our experiments identify \emph{mBART} as the
weakest pre-trained model, reporting the lowest accuracies for all
ATIS test languages. {\sc ZX-Parse} using \emph{XLM-R} generally
improved upon mBART for ATIS but proved worse for Overnight. As
\emph{XLM-R} is not pre-trained for sequence-to-sequence tasks, this
result suggests this model could be poorer at representing input
content in more complex queries. Despite being half the size of
\emph{XLM-R}, \emph{mBART50} is the only pre-trained encoder able to perform
competitively across all languages. Despite lower performance with different
pre-trained models, we identify that introducing additional objectives
yields improved accuracy in most cases. Similar to our results using \emph{mBART50},
we find that combining tasks is the optimal strategy for {\sc ZX-Parse} using
either \emph{XLM-R} or \emph{mBART} as an encoder. 

We additionally explored if pre-training
is required for our approach by training a comparable model from scratch.
While performance on English was similar to our best results, we found that
cross-lingual transfer was extremely poor and these results are omitted due
to negligible accuracies ($<2\%$) for non-English languages. Overall, this suggests
that our methodology is optimal when \emph{aligning an existing multi-lingual latent
space} rather than \emph{inducing a multi-lingual latent space from scratch}.

\paragraph*{Ablations of Reconstruction Language Data}

We present ablations to our main experiments examining the influence of
language similarity in reconstruction data for ATIS in Table
\ref{tab:lang_family_ablation_atis} and for Overnight in Table \ref{tab:lang_family_ablation_overnight}.
Similar to our results for Hindi and Turkish in Table \ref{tab:results_q2}, 
we find that using our auxiliary objectives in our model improves overall
cross-lingual alignment in languages that we did not intentionally target
with reconstruction data.

In our first case, we consider omitting the Romance genus languages (French, Spanish, Portuguese)
from the reconstruction corpus for experiments on ATIS. The observed reduction in
performance across all languages is likely a consequence of reduced training data
leading to weaker cross-lingual alignment. Notably, this drop is largest
for French ($-11.2\%$) and Spanish ($-7.1\%$). In contrast, the smallest reduction
is for Chinese ($-3.9\%$) and English ($-2.8\%$). We additionally examine the effect of 
omitting the only Sino-Tibetan language (Chinese) from experiments on
both ATIS and Overnight. While we observe a similar overall reduction in performance here
-- our notable finding is a larger reduction in parsing accuracy for Chinese
across both ATIS ($-17.0\%$) and Overnight ($-11.1\%$). Without a similar language
to Chinese (in the same family or with a similar orthography) in this experiment, we
suggest there is little to ``support'' better cross-lingual alignment for Chinese
relative to others. This contrasts with the performance drop for Romance languages,
which are still relatively similar to English and German. 

Overall, these ablations support that both \emph{variety} and \emph{similarity} are
important for considering language data for auxiliary objectives. Performance on
omitted languages can improve from a baseline, but better localization is achievable
using a linguistically varied ensemble of languages closely modeling the desired languages to parse. 


\end{document}